\documentclass[letterpaper]{article} 
\usepackage{aaai2026}  
\usepackage{times}  
\usepackage{helvet}  
\usepackage{courier}  
\usepackage[hyphens]{url}  
\usepackage{graphicx} 
\usepackage{natbib}  
\usepackage{caption} 
\usepackage{algorithm}
\usepackage{algorithmic}

\usepackage{newfloat}
\usepackage{listings}
\usepackage{pifont}
\usepackage{amsmath}
\usepackage{bm}
\DeclareCaptionStyle{ruled}{labelfont=normalfont,labelsep=colon,strut=off} 
\floatstyle{ruled}
\newfloat{listing}{tb}{lst}{}
\floatname{listing}{Listing}
 \nocopyright 

\title{MangaDiT: Reference-Guided Line Art Colorization \\
with Hierarchical Attention in Diffusion Transformers}
\author {
    Qianru Qiu\textsuperscript{\rm 1},
    Jiafeng Mao\textsuperscript{\rm 1},
    Kento Masui\textsuperscript{\rm 1},
    Xueting Wang\textsuperscript{\rm 1}
}
\affiliations {
    \textsuperscript{\rm 1}CyberAgent\\
    qiu\_qianru@cyberagent.co.jp, jiafeng\_mao@cyberagent.co.jp,\\ masui\_kento@cyberagent.co.jp,
    wang\_xueting@cyberagent.co.jp
}

\usepackage{bibentry}
\usepackage{booktabs}
\begin{document}

\maketitle

\begin{abstract}
Recent advances in diffusion models have significantly improved the performance of reference-guided line art colorization. However, existing methods still struggle with region-level color consistency, especially when the reference and target images differ in character pose or motion. Instead of relying on external matching annotations between the reference and target, we propose to discover semantic correspondences implicitly through internal attention mechanisms. In this paper, we present MangaDiT, a powerful model for reference-guided line art colorization based on Diffusion Transformers (DiT). Our model takes both line art and reference images as conditional inputs and introduces a hierarchical attention mechanism with a dynamic attention weighting strategy. This mechanism augments the vanilla attention with an additional context-aware path that leverages pooled spatial features, effectively expanding the model's receptive field and enhancing region-level color alignment. Experiments on two benchmark datasets demonstrate that our method significantly outperforms state-of-the-art approaches, achieving superior performance in both qualitative and quantitative evaluations.
\end{abstract}


\section{Introduction}

\begin{figure*}[t]
    \centering
    \includegraphics[width=1.0\linewidth]{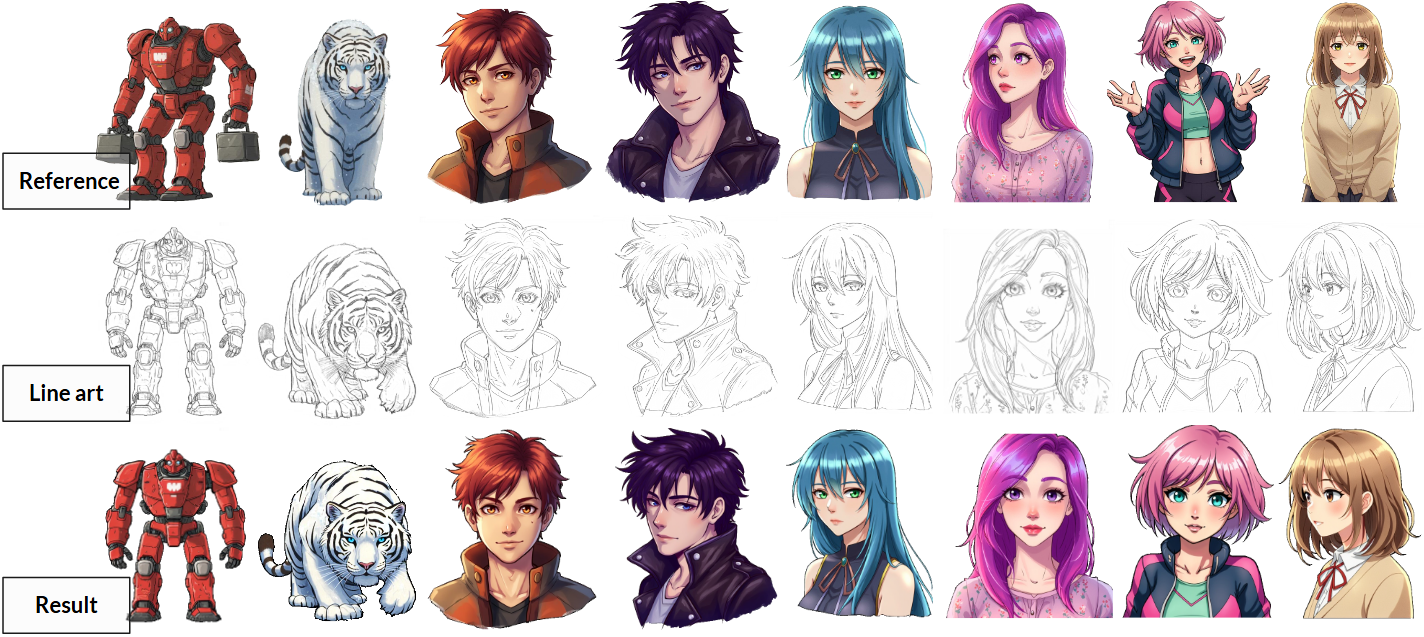}
    \caption{Reference-guided line art colorization results by MangaDiT. The samples are produced by image generation model.}
    \label{fig:overview}
\end{figure*}

The rapid advancements in generative models have revolutionized content creation in the anime and manga industries. Among these developments, reference-guided line art colorization, as shown in Figure~\ref{fig:overview}, has attracted substantial attention due to its practical value in creative workflows. This task enables efficient color transfer from a reference image to a line drawing, ensuring consistent coloring across corresponding semantic regions. 

However, reference-guided colorization remains challenging. The core issue lies in preserving color consistency, especially in regions with fine-grained details such as hair, clothing accessories, or intricate decorations. The task becomes even more difficult when character poses or motions differ significantly between the reference and target images. Existing methods~\cite{dai2024learning, yan2025image, yan2025colorizediffusion}, including diffusion-based generative models, have attempted to solve this issue but often struggle with maintaining region-level color alignment.
Other methods~\cite{liu2025manganinja, meng2025anidoc} aim to mitigate this challenge by incorporating external correspondence models to detect matching regions between the reference and target images. While effective in some cases, these methods rely heavily on the accuracy of external correspondence models. As these models are typically trained on natural images, they often lack a robust understanding of manga-style line art, making it difficult to accurately capture region-level correspondences. Consequently, incorrect region correspondence often leads to inconsistent color transfer.

We present MangaDiT, a powerful model for reference-guided line art colorization that leverages the strengths of Diffusion Transformers (DiT)~\cite{esser2024scaling}. Built upon DiT, our model effectively captures semantic connections between spatially separated regions by modeling long-range dependencies through transformer-based attention blocks. 
Unlike prior methods that rely on external correspondence annotations, MangaDiT learns to discover semantically aligned regions implicitly through its internal attention mechanisms. To further enhance region-level color consistency, we introduce a hierarchical attention mechanism with dynamic attention weighting. This design augments standard spatial attention with pooled contextual information, expanding the model’s receptive field and enabling more reliable color propagation across semantically similar regions.

MangaDiT's performance is evaluated on two benchmark datasets, one collected from animation videos with small character motions, and another newly created by us, specifically designed to test challenging cases with large character motions and significant pose variations. Experimental results demonstrate that our model outperforms existing state-of-the-art methods in both qualitative and quantitative assessments. We will release the code and benchmark dataset publicly upon acceptance of the paper.

In summary, our main contributions include:
\begin{itemize}
    \item Introduction of MangaDiT, a powerful DiT-based model for reference-guided line art colorization, trained via a lightweight LoRA-based fine-tuning strategy.
    \item Design of a hierarchical attention mechanism and a dynamic attention weighting strategy that enhance spatial attention with pooled contextual features, improving region-level color consistency.
    \item Demonstration of superior performance through extensive experiments on two benchmark datasets, including challenging scenarios with large character motion.
\end{itemize}
\section{Related Work}

\subsection{Reference-based Line Art Colorization}
Line art colorization aims to fill the blank regions of a line drawing with appropriate colors while preserving structural details and stylistic coherence. This task plays an important role in manga and anime production workflows. Various user-guided colorization methods have been explored, including text prompts~\cite{kim2019tag2pix}, scribbles~\cite{ci2018user, carrillo2023diffusart}, and reference images~\cite{yan2023two, cao2024animediffusion, cao2023attention}. Among them, reference-guided methods have become increasingly popular due to their intuitive and comprehensive nature of guidance. 
Recently, several works like BasicPBC~\cite{dai2024learning} and ColorizeDiffusion~\cite{yan2025image, yan2025colorizediffusion} have employed diffusion-based architectures for reference-guided colorization, leading to improved visual quality. However, these methods still struggle to achieve consistent region-level color transfer, particularly in the presence of character pose variations. To address this, some methods like MangaNinja~\cite{liu2025manganinja} and AniDoc~\cite{meng2025anidoc} attempt to improve alignment by integrating external point-to-point correspondence modules, such as LightGlue~\cite{lindenberger2023lightglue}, which detect matched keypoints between reference and target images. Nevertheless, these correspondence modules are typically pre-trained on natural image datasets and often perform suboptimally on manga-style line drawings due to domain gaps.

In contrast, our method avoids relying on such unreliable correspondences. Instead, we directly model region-level alignment through the internal attention mechanism of the network, enabling more robust and consistent colorization results.

\subsection{Conditional Diffusion Models}
Diffusion models have become a cornerstone of modern generative modeling in image synthesis, offering strong flexibility to incorporate various conditional inputs during the denoising process. Early conditional diffusion models, such as Stable Diffusion~\cite{rombach2022high}, introduce a latent-space formulation where images are generated in a compressed latent space. These models typically use a U-Net backbone as the denoising network, with cross-attention layers enabling conditioning on text prompts. To further enhance controllability, ControlNet~\cite{zhang2023adding} is proposed to inject structural conditions by training parallel control branches within the U-Net. Recently, Diffusion Transformers (DiT)~\cite{esser2024scaling} replace the U-Net backbone with pure transformer blocks, achieving superior performance in modeling long-range semantic dependencies.
Building on this, OminiControl~\cite{tan2025ominicontrol} explores how to integrate diverse conditional signals into DiT-based architectures. It systematically compared two paradigms: ControlNet-style structural branching and parameter-efficient fine-tuning methods such as LoRA~\cite{hu2022lora}. Their results show that in transformer-based diffusion models, LoRA-based tuning achieves comparable controllability with significantly lower computational overhead. 

Inspired by these insights, we build upon the DiT models and adopts a LoRA-based fine-tuning strategy to enable efficient training for reference-guided line art colorization. 

\section{Approach}

\begin{figure*}[t]
    \centering
    \includegraphics[width=1.0\linewidth]{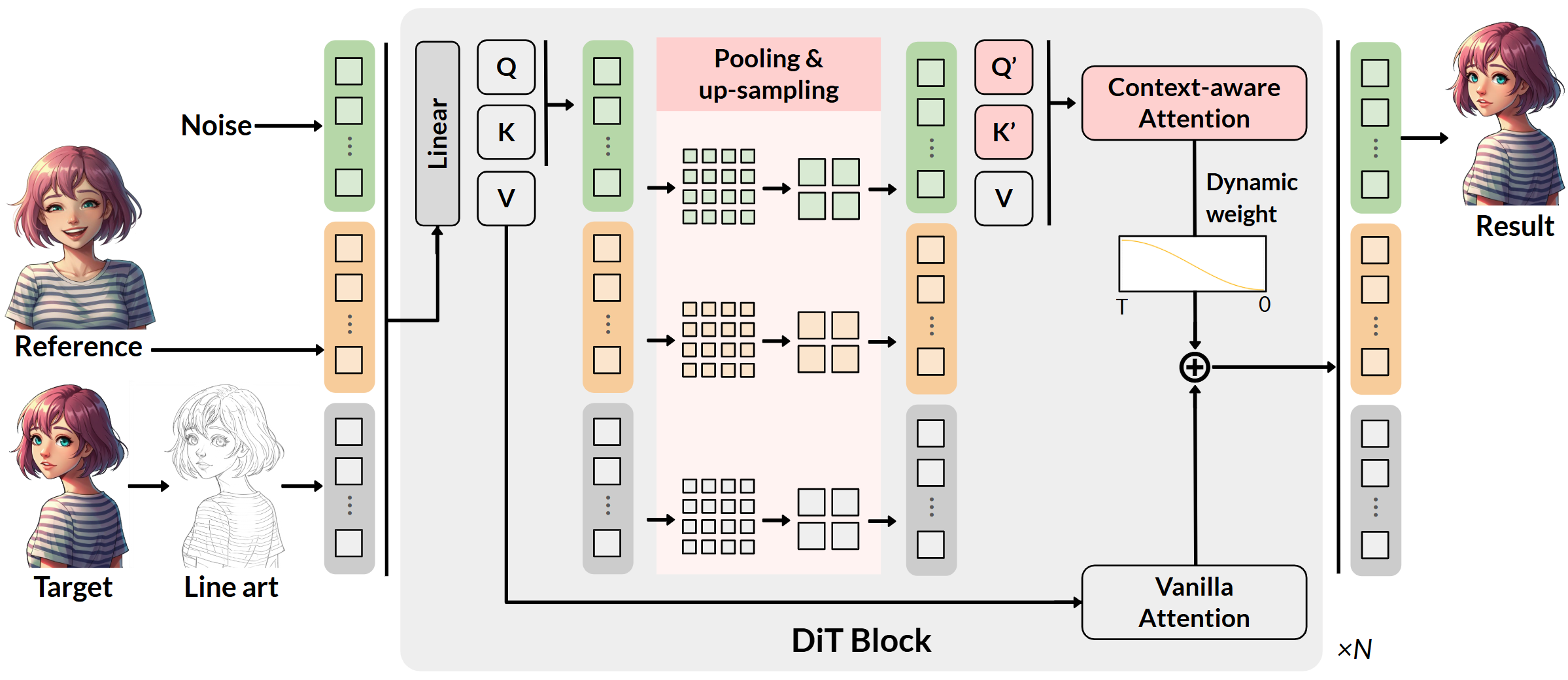}
    \caption{Overview of reference-guided line art colorization with hierarchical attention in DiT block. The empty text condition is omitted here. The modified query and key vectors ($Q'$ and $K'$) for context-aware attention are processed through the same pooling and upsampling procedure for each image features, which preserving the original token structure. The resulting context-aware attention is added to the vanilla attention with a dynamic weight scheduled over diffusion timesteps.}
    \label{fig:framework}
\end{figure*}

\subsection{Preliminary}
Diffusion models operate in two stages: a forward noising process and a reverse denoising process. In the forward process, noise is gradually added to a target image $x_0$ over a series of time steps $t \in [1, T]$, where T denotes the total number of steps. The denoising process then aims to progressively remove the noise from a noisy image $x_T$ to generate a clean image output. The DiT model, used in architectures such as FLUX.1~\cite{flux.1} and Stable Diffusion 3~\cite{esser2024scaling}, employs transformer blocks as the denoising network to iteratively refine noisy image tokens. 
During training, the target image and the text prompt are first encoded into latent tokens using frozen encoders. The noisy image tokens $X$ are initialized from Gaussian noise in the latent space. The DiT model takes the noisy image tokens $X$ and the text condition tokens $C_{T}$ as input, and learns to denoise the image step by step through a reverse diffusion process. 

The spatial representation of the image, with dimensions $N \times N$, is flattened into a sequence of $N^2$ tokens, which are embedded and processed by multiple transformer blocks. Each DiT block consists of layer normalization followed by multi-modal attention to capture spatial information. The multi-modal attention mechanism projects the position-encoded tokens into query $Q$, key $K$, and value $V$ representations, allowing attention computation across all tokens of multiple conditional inputs.

Due to its modular design, the DiT architecture is highly flexible and can be extended to incorporate various types of input conditions.

\subsection{Conditioning Integration Strategy}
We extend the DiT architecture to support multiple visual conditions for reference-guided line art colorization. As shown in Figure~\ref{fig:framework}, our model takes three types of visual inputs: the noisy latent representation for the target image, its target line art, and a colored reference image. To integrate these conditions, we adopt a unified token sequence design. Specifically, the latent tokens from the noisy image $X$, the text prompt condition $C_T$, the line art condition $C_L$, and the reference image condition $C_R$ are concatenated into a single sequence $[X, C_T, C_L, C_R]$. This unified token sequence is then passed through the DiT blocks using standard multi-head self-attention, enabling information exchange across all tokens without requiring architectural modifications. The vanilla attention mechanism is computed as:
\begin{equation}
\mathrm{A_{vanilla}}([X, C_{T}, C_{L}, C_{R}]) = \mathrm{softmax}\left( \frac{Q K^\top}{\sqrt{d_k}} \right) V
\end{equation}

To efficiently adapt the pre-trained DiT model to our colorization task, we follow a parameter-efficient fine-tuning strategy by applying LoRA to the attention layers. This allows the model to learn task-specific conditioning behaviors without modifying the full set of transformer weights. Based on this setup, our model learns to discover semantic correspondences between structural features in the line art and appearance cues in the reference image.

\subsection{Hierarchical Attention Mechanism}
While the DiT-based baseline can transfer global color information from the reference image to the target line art, we observe that it often fails to maintain local color consistency in semantically corresponding regions, especially in detailed areas such as clothing or accessories. We attribute this limitation to the restricted receptive field of standard self-attention, which focuses on token-wise interactions and lacks explicit access to broader spatial context. 
To address this issue, we introduce a hierarchical attention mechanism that augments vanilla spatial attention with an additional context-aware attention path derived from pooled feature representations. This design enables the model to integrate coarse-level contextual cues, improving its ability to propagate colors consistently across semantically related regions. Specifically, we first extract token sequences for the noisy image, the line art image, and the reference image, then reshape each into a spatial feature map of size $N \times N$. We apply max pooling with a randomly selected kernel size from $[2, 4, 8]$ to capture coarse semantics at multiple scales. These pooled features are then upsampled back to the original resolution via nearest-neighbor interpolation. These upsampled features are projected into a separate set of query and key representations, $Q'$ and $K'$, and used to compute the context-aware attention:
\begin{equation}
\mathrm{A_{context}}([X, C_{T}, C_{L}, C_{R}]) = \mathrm{softmax}\left( \frac{Q' K'^\top}{\sqrt{d_k}} \right) V
\end{equation}

The final hierarchical attention $\mathrm{A_{hier}}$ is obtained by blending the vanilla and context-aware attention with a dynamic weight $\lambda$:
\begin{equation}
\mathrm{A_{hier}} = \mathrm{A_{vanilla}} + \lambda \mathrm{A_{context}}
\end{equation}

This hierarchical attention is applied throughout the denoising process, with its influence modulated over time using a timestep-dependent weighting strategy.

\subsection{Dynamic Weight with Cosine Scheduling}

Although the hierarchical attention mechanism enhances region-level color consistency by introducing coarse contextual features, its contribution should not remain constant throughout the denoising process. Prior works~\cite{choi2022perception, lin2024tasr, cho2024enhanced} have shown that using timestep-dependent control weights in diffusion models leads to more stable and effective generation. In the denoising process, earlier timesteps (large $t$) correspond to high-noise states, where the model focuses on coarse structure generation. Later timesteps (small $t$) focus on refining fine-grained details. Therefore, the importance of coarse context should be emphasized more in early stages and gradually reduced as the model progresses toward detailed refinement.
To reflect this intuition, we introduce a dynamic attention weighting strategy that adjusts the strength of the hierarchical attention over time. We define a timestep-dependent weight $\lambda(t)$ using a cosine schedule:
\begin{equation}
\mathrm{\lambda}(t) = \lambda_{base} \times 0.5 \times (1 - \cos(\frac{\pi t}{T}))
\end{equation}

Here, $\lambda_{base}$ is the maximum blending weight (set to 0.1 in our experiments). This scheduling ensures that the influence of coarse-level attention diminishes at later timesteps, allowing the model to prioritize fine-grained detail preservation. We further investigate the effectiveness of the dynamic weighting strategy, and compare different weight scheduling in ablation study.


\section{Experiments}

\subsection{Datasets}

For the fine-tuning phase, we utilize the sakuga-42m dataset, which has been used in prior works~\cite{liu2025manganinja, meng2025anidoc}. We extract keyframes from this dataset and construct training pairs by initially setting the frame interval between the reference and target frames to 18. If the number of matching keypoints, determined by LightGlue~\cite{lindenberger2023lightglue}, is fewer than 25, we iteratively reduce the interval until a pair with sufficient correspondence is found. The line art images are estimated using an off-the-shell LineartAnimeDetector model~\cite{zhang2023adding}
. The final dataset consists
of reference, target, and line art images. No text prompts are used in this work.

For the evaluation phase, 
since related works~\cite{yan2025image, yan2025colorizediffusion, liu2025manganinja} do not publicly release their evaluation datasets,
we prepare two new benchmark datasets to comprehensively assess model performance. The first evaluation set is selected from the public ATD-12K~\cite{siyao2021deep}, where each sample contains three keyframe images. We select two of them as the reference-target pair. Line art images are generated using the same method as the training set. We select 200 triplets of reference, target, and line art images to form the ATD-test200 dataset. These samples include both foreground characters and background elements, with relatively small pose differences between frames. In addition, we manually segment each image to extract the foreground subjects, producing a variant we denote as ATD-test200-fg. 
To evaluate performance under more challenging motion variations, we construct a synthetic dataset using Unity and VRoid Studio. We manually create 20 distinct 3D characters and render each character in multiple poses, resulting in 200 reference-target image pairs. For each pair, we generate the corresponding line art images
and remove background elements to isolate the character, forming the Unity-test200 dataset. More details about this dataset are provided in the supplementary materials. We plan to release this dataset publicly upon acceptance of the paper.

Due to copyright constraints in ATD-test200, we create several reference–target pairs generated by a text-to-image model, used solely in figures for visualization. Further details are available in the supplementary materials.

\subsection{Experimental Setup}

\textbf{Implementation Details.} 
Our model is built upon FLUX.1-dev~\cite{flux.1}, a latent rectified flow transformer model. To adapt it for reference-guided line art colorization, we apply LoRA to the attention layers, using a default rank of 4. All experiments are conducted on a single NVIDIA A100-80GB GPU for 50,000 iterations. Training takes about 36 hours with a batch size of 1 and gradient accumulation over 8 steps. For evaluation, we compare both the final checkpoint at step 50,000 and the Exponential Moving Average (EMA) version selected based on evaluation performance, and report results from the better-performing model.

\par\addvspace{0.5em}
\noindent\textbf{Compared methods.} 
We compare our method with four recent state-of-the-art approaches for reference-guided line art colorization: BasicPBC~\cite{dai2024learning}, ColorizeDiffusion v1.0~\cite{yan2025colorizediffusion} (denoted as ColDiff1.0), ColorizeDiffusion v1.5~\cite{yan2025image}(denoted as ColDiff1.5), and MangaNinja~\cite{liu2025manganinja}. All methods are evaluated using the same line art and reference image pairs with a fixed noise seed to ensure fair comparison.

\par
\begin{figure}[t]
    \centering
    \includegraphics[width=1.0\linewidth]{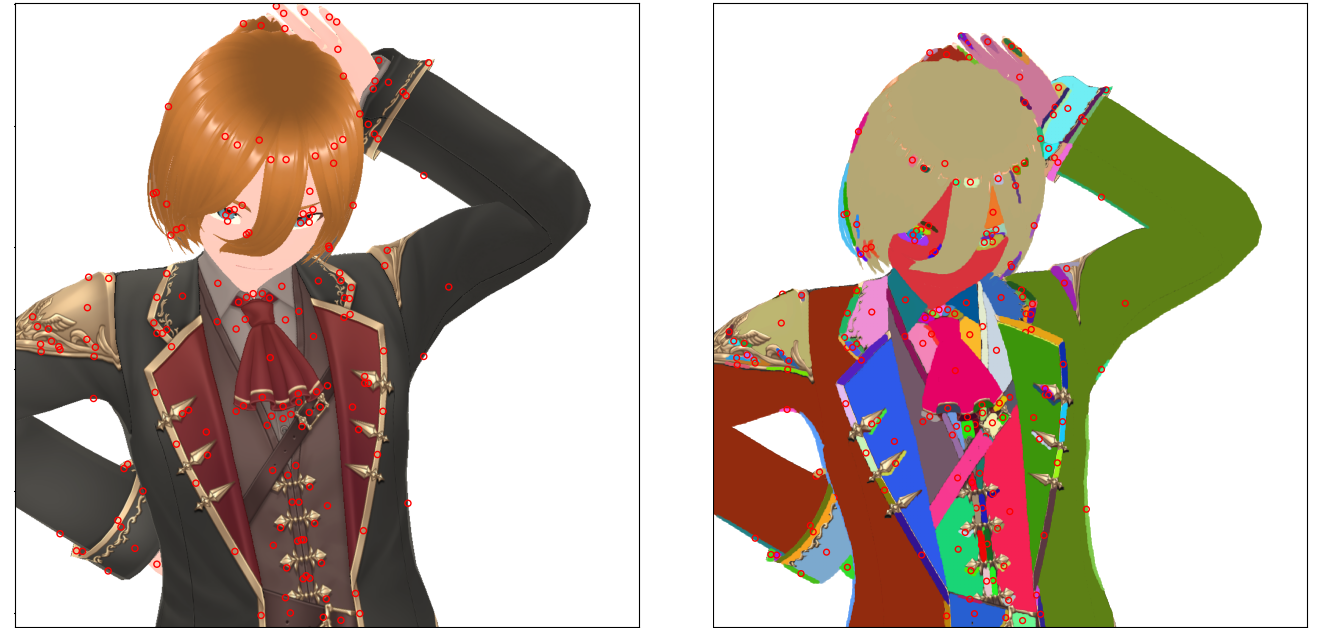}
    \caption{A sample of segmented color regions. The left image is the ground truth image and the right image is the segmented output with color regions. The segmented regions are denoted with different colors and the red point is the represented pixel in the color region. }
    \label{fig:color_regions}
\end{figure}

\addvspace{0.5em}
\noindent\textbf{Evaluation Metrics.}
To evaluate the quality of the generated images, we adopt the following metrics: CLIP~\cite{shen2022k} semantic image similarities (CLIP), Peak Signal-to-Noise Ratio (PSNR)~\cite{wang2004image}, Structural Similarity Index Measure (SSIM)~\cite{huynh2008scope}, and Learned Perceptual Image Patch Similarity (LPIPS)~\cite{zhang2018unreasonable}. In addition to these standard metrics, we introduce a region-based point-wise color error metric, denoted as Mean Squared Error of Color Regions ($\text{MSE}_\text{CR}$), to more accurately assess color consistency. This metric emphasizes local color fidelity. To compute $\text{MSE}_\text{CR}$, we first segment the ground truth image into fine-grained color regions based on CIELab color differences. As illustrated in Figure~\ref{fig:color_regions}, the segmentation separates even semantically similar regions if they are divided by line boundaries, ensuring precise region-level evaluation. Each segmented region is assigned a representative pixel, and the squared color difference between the corresponding pixels in the generated and ground truth images is calculated. 

\section{Results}

\begin{figure*}[t]
    \centering
    \includegraphics[width=1.0\linewidth]{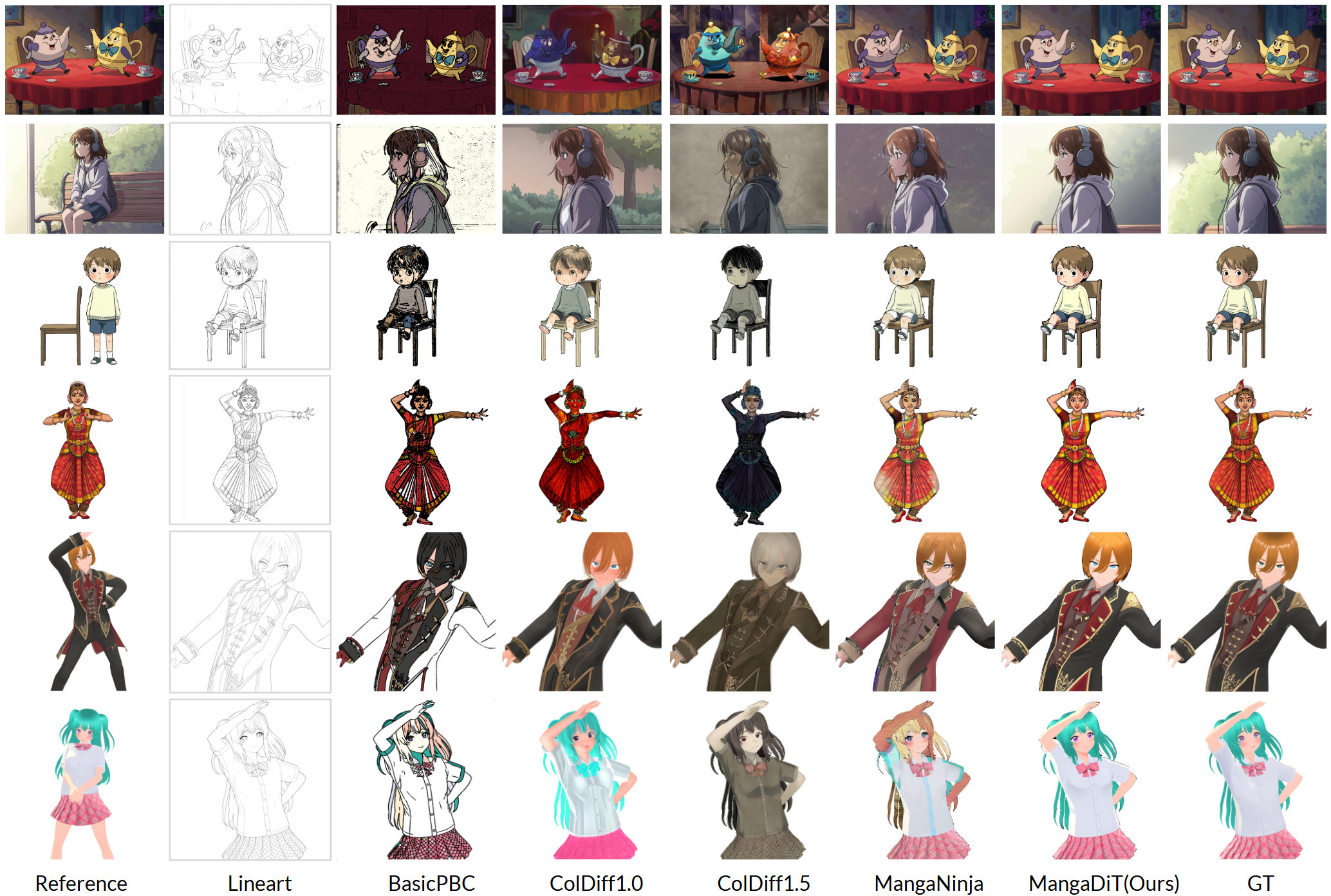}
    \caption{Qualitative comparison of line art colorization results across different methods. The first four rows show samples generated by a text-to-image model, where the top two include backgrounds and the middle two are foreground-only. The final two rows display samples from the Unity-test200 dataset.}
    \label{fig:colorization_results}
\end{figure*}

\begin{table}[t]
    \caption{Quantitative results on ATD-test200 dataset}
    \label{tab:res_atd}
    \setlength\tabcolsep{2pt}
    \hbox to\hsize{\hfil
    \begin{tabular}{l|c|c|c|c|c}\hline
        \toprule
        Method & {CLIP$\uparrow$} & {PSNR$\uparrow$} & {SSIM$\uparrow$} & {LPIPS$\downarrow$} & $\text{MSE}_\text{CR}\downarrow$ \\
        \midrule
        BasicPBC &  $0.755$ & $12.7$ & $0.227$ & $0.591$ & $0.119$ \\
        ColDiff1.0 & $0.900$ & $13.0$ & $0.293$ & $0.302$ & $0.095$ \\
        ColDiff1.5 &  $0.722$ & $8.1$ & $0.137$ & $0.3800$ & $0.241$ \\
        MangaNinja & $0.912$ & $13.9$ & $0.511$ & $0.250$ & $0.103$ \\
        \midrule
        $\text{MangaDiT}_\text{(Ours)}$ & $\bm{0.965}$ & $\bm{27.8}$ & $\bm{0.944}$ & $\bm{0.059}$ & $\bm{0.004}$ \\
        \bottomrule
    \end{tabular}\hfil}
\end{table}

\begin{table}[t]
    \caption{Quantitative results on ATD-test200-fg dataset}
    \label{tab:res_atd_fg}
    \setlength\tabcolsep{2pt}
    \hbox to\hsize{\hfil
    \begin{tabular}{l|c|c|c|c|c}\hline
        \toprule
        Method & {CLIP$\uparrow$} & {PSNR$\uparrow$} & {SSIM$\uparrow$} & {LPIPS$\downarrow$} & $\text{MSE}_\text{CR}\downarrow$ \\
        \midrule
        BasicPBC & $0.825$ & $5.86$ & $0.218$ & $0.560$ & $0.329$ \\
        ColDiff1.0 & $0.870$ & $12.9$ & $0.329$ & $0.246$ & $0.125$ \\
        ColDiff1.5 & $0.837$ & $9.1$ & $0.188$ & $0.331$ & $0.169$ \\
        MangaNinja & $0.876$ & $14.1$ & $0.518$ & $0.201$ & $0.080$ \\
        \midrule
        $\text{MangaDiT}_\text{(Ours)}$ & $\bm{0.951}$ & $\bm{22.0}$ & $\bm{0.844}$ & $\bm{0.085}$ & $\bm{0.011}$ \\
        \bottomrule
    \end{tabular}\hfil}
\end{table}

\begin{table}[t]
    \caption{Quantitative results on Unity-test200 dataset}
    \label{tab:res_unity}
    \setlength\tabcolsep{2pt}
    \renewcommand{\arraystretch}{1.2}
    \hbox to\hsize{\hfil
    \begin{tabular}{l|c|c|c|c|c}\hline
        \toprule
        Method & {CLIP$\uparrow$} & {PSNR$\uparrow$} & {SSIM$\uparrow$} & {LPIPS$\downarrow$} & $\text{MSE}_\text{CR}\downarrow$ \\
        \midrule
        BasicPBC & $0.883$ & $8.7$ & $0.228$ & $0.392$ & $0.217$ \\
        ColDiff1.0 & $0.936$ & $13.8$ & $0.403$ & $0.226$ & $0.102$ \\
        ColDiff1.5 & $0.837$ & $9.1$ & $0.188$ & $0.331$ & $0.169$ \\
        MangaNinja & $0.896$ & $11.3$ & $0.327$ & $0.277$ & $0.131$ \\
        \midrule
        $\text{MangaDiT}_\text{(Ours)}$ & $\bm{0.944}$ & $\bm{17.3}$ & $\bm{0.655}$ & $\bm{0.163}$ & $\bm{0.066}$ \\
        \bottomrule
    \end{tabular}\hfil}
\end{table}

\begin{figure}[t]
    \centering
    \includegraphics[width=1.0\linewidth]{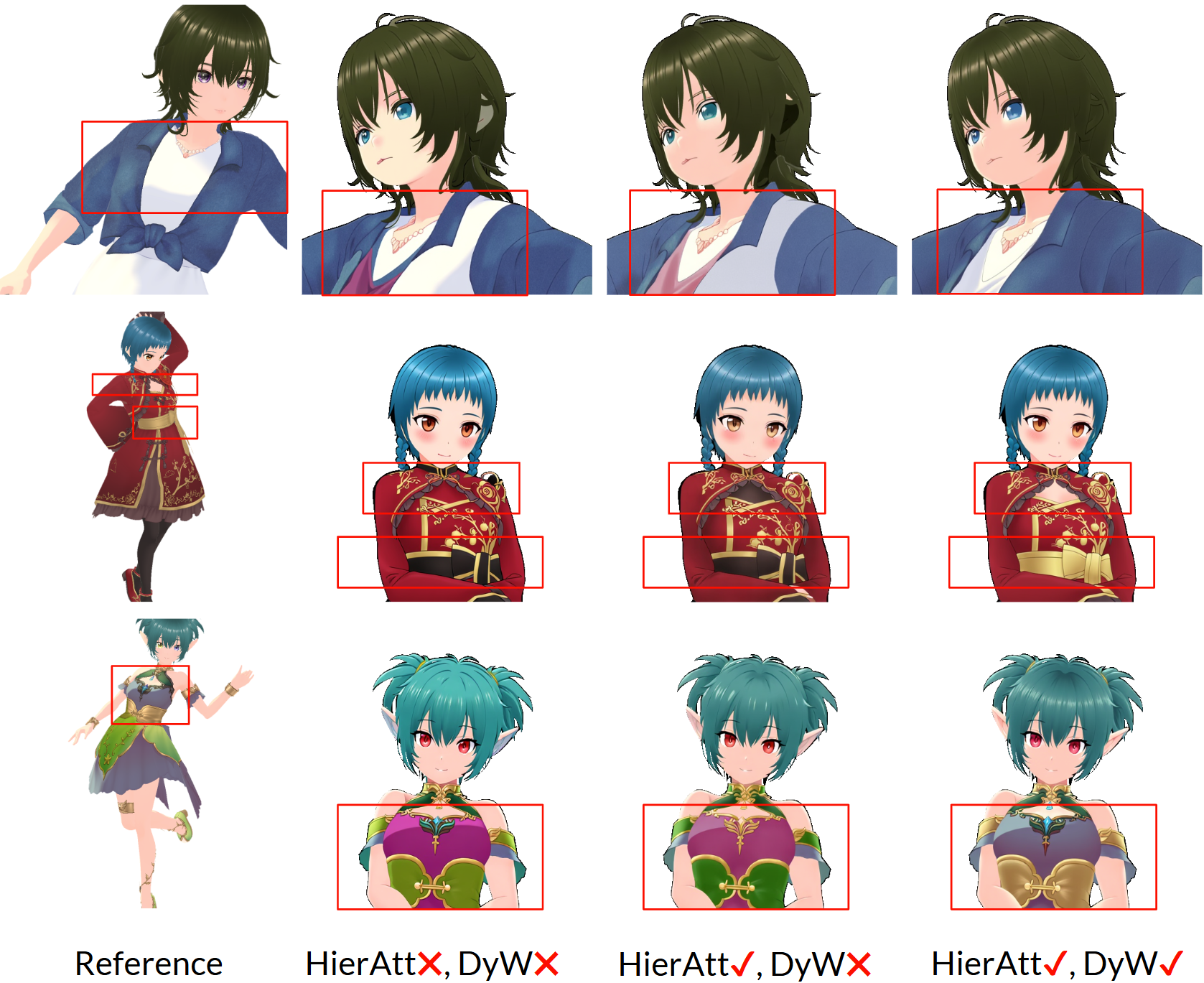}
    \caption{Qualitative comparison of our models with different training strategies.}
    \label{fig:results_ablation}
\end{figure}

\begin{table}[t] 
    \caption{Ablation results of different training strategies on Unity-test200 dataset}
    \label{tab:res_ablation}
    \setlength\tabcolsep{2.5pt}
    \renewcommand{\arraystretch}{1.2}
    \hbox to\hsize{\hfil
    \begin{tabular}{c|c|c|c|c|c|c}\hline
        \toprule
        HierAtt & DyW & CLIP$\uparrow$ & {PSNR$\uparrow$} & {SSIM$\uparrow$} & {LPIPS$\downarrow$} & $\text{MSE}_\text{CR}\downarrow$ \\
        \midrule
        \ding{55} & \ding{55} & $0.939$ & $15.8$ & $0.593$ & $0.181$ & $0.083$ \\
        \ding{51} & \ding{55} & $0.938$ & $16.6$ & $0.621$ & $0.176$ & $0.072$ \\
        \ding{51} & \ding{51} & $\bm{0.944}$ & $\bm{17.3}$ & $\bm{0.655}$ & $\bm{0.163}$ & $\bm{0.066}$ \\
        \bottomrule
    \end{tabular}\hfil}
\end{table}

\subsection{Qualitative and Quantitative Results}

We evaluate our method on three variants of two datasets, ATD-test200, ATD-test200-fg, and Unity-test200, to assess both colorization accuracy and robustness under varying degrees of character motion between the reference and target images. The quantitative results of these datasets are separately presented in Table~\ref{tab:res_atd}, Table~\ref{tab:res_atd_fg}, and Table~\ref{tab:res_unity}. Across all benchmarks and evaluation metrics, our method consistently outperforms existing approaches, demonstrating superior performance in scenarios involving both minor and significant pose or motion variations.

The qualitative results are shown in Figure~\ref{fig:colorization_results}, showcasing the results of our full model. As illustrated, our method yields higher-quality foreground colorization, particularly in regions with complex structures. Moreover, prior methods often fail to generate coherent background content that aligns well with the reference image. In contrast, our approach is capable of generating backgrounds that closely match the reference image, substantially enhancing visual consistency.
This substantially improves the practical utility of our model in real-world applications, where both foreground fidelity and background coherence are crucial.

\subsection{Ablation Study} \label{sec:ablation}

\begin{table}[t] 
    \caption{Ablation results of different weight scheduling on Unity-test200 dataset}
    \label{tab:res_weight}
    \setlength\tabcolsep{3.5pt}
    \renewcommand{\arraystretch}{1.2}
    \hbox to\hsize{\hfil
    \begin{tabular}{l|c|c|c|c|c}\hline
        \toprule
        Schedule & {CLIP$\uparrow$} & {PSNR$\uparrow$} & {SSIM$\uparrow$} & {LPIPS$\downarrow$} & $\text{MSE}_\text{CR}\downarrow$ \\
        \midrule
        sin & $0.936$ & $15.7$ & $0.585$ & $0.185$ & $0.084$ \\
        cosInv & $0.939$ & $16.8$ & $0.629$ & $0.169$ & $0.071$ \\
        cos (ours) & $\bm{0.944}$ & $\bm{17.3}$ & $\bm{0.655}$ & $\bm{0.163}$ & $\bm{0.066}$ \\
        \bottomrule
    \end{tabular}\hfil}
\end{table}

\textbf{Ablation of training strategies.} 
We conduct a series of ablation experiments to examine how different components of our framework contribute to overall colorization performance, with a particular focus on region-level color consistency. We evaluate three model variants that differ in their use of the hierarchical attention mechanism (HierAtt) and dynamic weighting with cosine scheduling (DyW). In the first setting, only the reference image and line art are used as conditional inputs to the attention mechanism, and the hierarchical attention module is disabled. In the second setting, we enable the hierarchical attention mechanism by introducing pooled context-aware attention, but apply a constant attention weight across all diffusion timesteps. In the third setting, we integrate context-aware attention with dynamic attention weighting using a cosine schedule, allowing the influence of hierarchical attention to gradually decrease as the diffusion process progresses. To better isolate the impact of these strategies, we conduct our analysis on the Unity-test200 benchmark, where the reference and target images exhibit significant pose and motion differences. 
The quantitative results are reported in Table~\ref{tab:res_ablation}, and visual comparisons are shown in Figure~\ref{fig:results_ablation}. We observe that, enabling hierarchical attention yields noticeable improvements, and further applying dynamic attention weighting leads to enhanced region-level color consistency in the generated results.

\par\addvspace{0.5em}
\noindent\textbf{Ablation of weighting strategies.}
We further compare different weight scheduling methods for the dynamic attention weight to analyze how the modulation of hierarchical attention over timesteps affects performance. Specifically, we evaluate three weighting strategies: (1) sinusoidal schedule (sin), where the weight is low at both the beginning and end of the diffusion process and peaks in the middle, computed as $\lambda_{base} \times \sin(\frac{\pi t}{T})$; (2) inverse cosine schedule (cosInv), where the weight is high at early timesteps and gradually decreases, computed as $\lambda_{base} \times 0.5 \times (1 + \cos(\frac{\pi t}{T}))$; (3) our proposed cosine schedule (cos), where the weight is high at large timesteps and decreases over time. As shown in Table~\ref{tab:res_weight}, the cosine schedule achieves the best performance across all evaluation metrics. This confirms our hypothesis that applying stronger hierarchical guidance in the early stages of denoising, when the model primarily focuses on global structure, helps establish better region-level correspondence. As the denoising progresses toward fine detail refinement, reducing the influence of coarse context allows the model to focus more on localized appearance features and achieve higher image quality. Additional ablation studies on attention integration variants and base weight settings are provided in the supplementary materials.

\begin{figure}[t]
    \centering
    \includegraphics[width=1.0\linewidth]{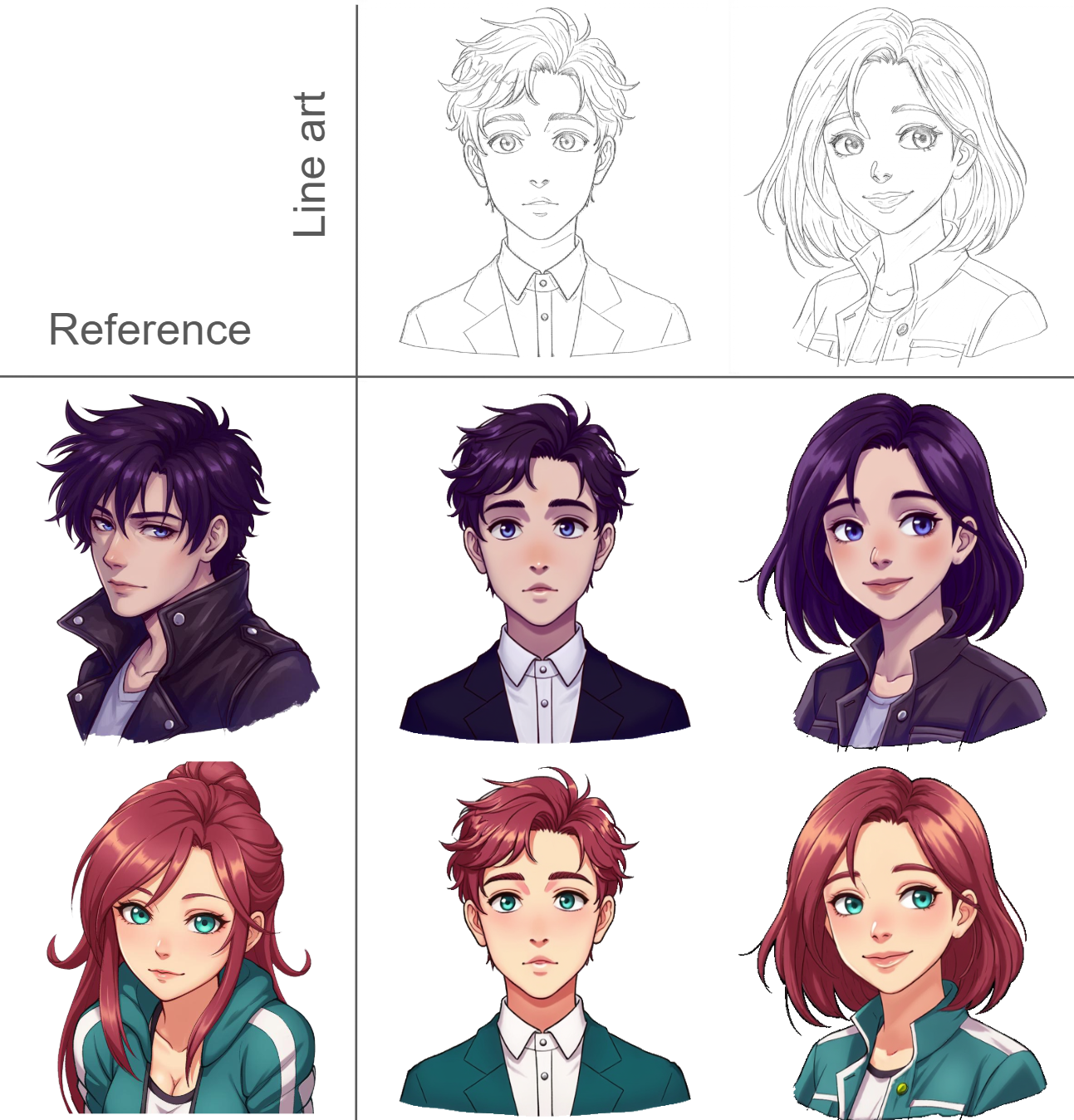}
    \caption{Colorization results with  references of different characters.}
    \label{fig:cross_char}
\end{figure}

\begin{figure}[t]
    \centering
    \includegraphics[width=1.0\linewidth]{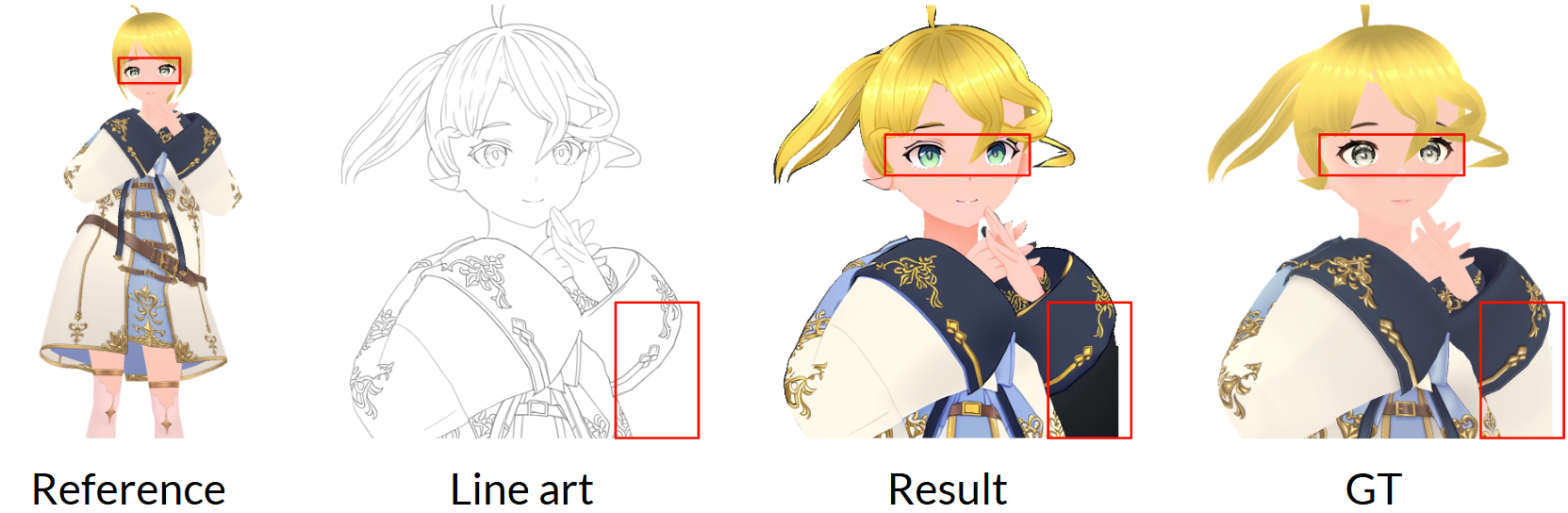}
    \caption{Error case with region limited line art.}
    \label{fig:error_case1}
\end{figure}

\subsection{Colorization with References of Different Characters}

Although our method is primarily designed for reference-guided colorization in which the reference and target line art depict the same character, we observe that the model can still generate reasonable results even when the reference image shows a different character. In such cases, the model transfers general color patterns, such as eye, hair, and clothing colors, while adapting them to fit the structural characteristics of the target line art. As shown in Figure~\ref{fig:cross_char}, the model demonstrates robustness in handling appearance discrepancies and produces coherent, visually plausible colorization results.

\subsection{Limitation}

While our method achieves strong performance in reference-guided line art colorization, it is important to acknowledge the limitation arising from the absence of line structures. The model may struggle in regions where the line art does not clearly convey the underlying semantics. As shown in Figure~\ref{fig:error_case1}, when the line drawing of a sleeve is incomplete, the model may fail to align this region with the corresponding area in the reference image, resulting in color mismatches. In addition, when the reference image contains very small regions, the model may produce inaccurate colors in the corresponding semantic areas of the target image. These ambiguities can cause errors in region matching and colorization. To mitigate such issues, we recommend providing detailed and complete line art, as well as reference images with sufficient and clearly visual information.
\section{Conclusion}

In this work, we proposed a novel approach for reference-guided line art colorization using the Diffusion Transformer (DiT) architecture. By introducing hierarchical attention and dynamic attention weighting, our model effectively improves region-level color consistency, particularly under significant pose or motion variations between reference and target images. Experiments on two benchmark datasets show that our method outperforms previous state-of-the-art approaches in both quantitative and qualitative evaluations.
Overall, the proposed method shows strong potential for improving the quality and efficiency of reference-guided colorization workflows, making it a practical and effective tool for digital artists and animators.

\bibliography{aaai2026}

\end{document}